\documentclass[]{interact}

\usepackage{epstopdf}

\usepackage{natbib}
\bibpunct[, ]{(}{)}{;}{a}{}{,}
\renewcommand\bibfont{\fontsize{10}{12}\selectfont}

\theoremstyle{plain}

\theoremstyle{definition}

\theoremstyle{remark}

\usepackage{subfig}
\usepackage{algorithm,algpseudocode}
\usepackage{url}

\begin{document}

\articletype{IJCAI 2023 Workshop on Search and Planning with Complex Objectives (Oral Presentation)}

\title{Using Reinforcement Learning for the Three-Dimensional Loading Capacitated Vehicle Routing Problem}

\author{
\name{Schoepf S.\textsuperscript{a}\thanks{Schoepf S. Email: ss2823@cam.ac.uk}, Mak S.\textsuperscript{a}, Xu L.\textsuperscript{a}, Senoner J.\textsuperscript{b}, Netland T.\textsuperscript{b}, Brintrup A.\textsuperscript{a}}
\affil{\textsuperscript{a} Department of Engineering, University of Cambridge, Cambridge, United Kingdom \\
\textsuperscript{b} Department of Management, Technology and Economics, ETH Zurich, Zurich, Switzerland}
}

\maketitle

\begin{abstract}
HGVs are important for the economy but contribute significantly to climate change with only 60\% loading efficiency in the UK. Collaborative vehicle routing with co-loading of delivery items is a promising solution to increase efficiency, but challenges remain to make this a possibility. One key challenge is the efficient computation of viable solutions. Current operations research methods suffer from non-linear scaling with increasing problem size and are therefore bound to limited geographic areas to compute results in time for day-to-day operations. This only allows for local optima in routing and leaves global optimization potential untouched. We develop a reinforcement learning model to solve the three-dimensional loading capacitated vehicle routing problem in linear time. While the three-dimensional loading capacitated vehicle routing problem has been studied extensively in operations research, no publications on solving the problem with reinforcement learning exist. We demonstrate the linear time scaling of our reinforcement learning model and benchmark our routing performance against state-of-the-art methods. The model performs within an average gap of 3.83\% to 7.65\% compared to the established methods. Our model, therefore, not only represents a promising first step towards large scale logistics optimization with reinforcement learning but also lays the foundation for this stream of research. GitHub: https://github.com/if-loops/3L-CVRP
\end{abstract}


\section{Introduction}\label{sec:intro}
Heavy goods vehicles (HGVs) are a vital part of the supply chains that power our economy. At the same time, HGVs are also a major contributor to climate change, accounting for 4.75\% of the total greenhouse gas emissions in the United Kingdom alone \citep{greenhouse}. 
Despite this significant impact on our environment, logistics operations in the UK are lacking efficiency with an average loading factor of only around 60\% for HGVs \citep*{fta}. 

\begin{figure}[h]
  \centering
  \includegraphics[width=0.7\textwidth]{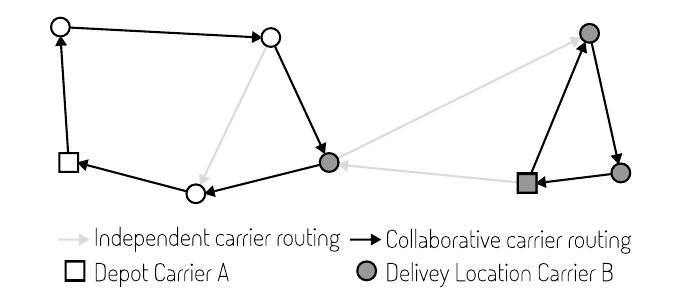}
  \caption{Collaborative packing and routing}
  \label{fig:collab}
\end{figure}

One promising approach to increase the efficiency of HGV usage is collaborative vehicle routing with co-loading of delivery items, as shown in Figure \ref{fig:collab}. In this approach, carriers give away packages that are undesirable for their routes and accept packages from other carriers that fit within their routes. This exchange of items between carriers enables more efficient operations than a single carrier could achieve individually. The problem in practice lies within the sharing and assignment of the transportation requests \citep{collabrouting}. While there are different centralized and decentralized approaches to this challenge, they all rely on bin-packing and vehicle-routing algorithms to determine viable package reassignment options. The combined problem of vehicle routing and 3D bin-packing is commonly referred to as the three-dimensional loading capacitated vehicle routing problem (3L-CVRP).

\begin{figure}[h]
  \centering
  \includegraphics[width=0.7\textwidth]{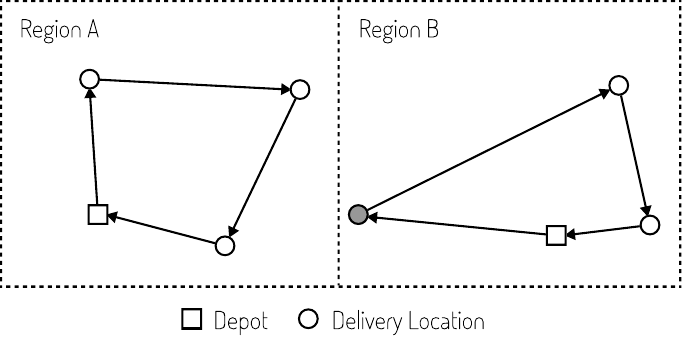}
  \caption{Optimality gap caused by regional optimization}
  \label{fig:regions}
\end{figure}

The 3L-CVRP is a long-standing problem in the operations research literature (e.g., {Gendreau et al. 2006}). However, these existing approaches have two shortcomings.
First, existing approaches are based on heuristics that are limited by the quality of the chosen rules and need to be adapted to new scenarios by domain experts. Second, existing approaches require convergence towards near-optimal solutions, which is computationally expensive. For example, a routing problem with 15 destinations requires around 10 seconds, while a routing problem with 100 destinations requires over 2000 seconds \citep{mahvashcolumn}.
In practice this leads to the decomposition into different regions (e.g., area codes) that get optimized individually to enable computation on a day-to-day basis. As illustrated in Figure \ref{fig:regions} compared to Figure \ref{fig:collab}, this approach creates solutions that can be locally optimal but lack global optimality due to the sub-optimal decomposition.

Solving the shortcomings of existing methods in the 3L-CVRP requires a method that relies less on heuristics and significantly improves compute times. For this, we propose the use of reinforcement learning (RL). 
Compared to existing approaches, reinforcement learning has several benefits. First, while training a reinforcement learning model is time-intensive, the learned policies can be executed in near real-time. Second, reinforcement learning models can learn policies based on a given cost function without human-designed heuristics. This not only allows us to potentially close the global optimality gap shown in Figure \ref{fig:regions} but also enables retraining for new problem settings without the need for new expert generated heuristics.

We benchmark the proposed reinforcement learning model on the original formulation of the 3L-CVRP (i.e., Gendreau et al. 2006) against state-of-the-art approaches \citep[e.g.,][]{mahvashcolumn, zhang_evolutionary_2015}.
Our benchmarking results show an average gap of 3.83\% to \citet{mahvashcolumn} and 7.65\% to \citet{zhang_evolutionary_2015} combined with significantly better compute time scaling. Our model, therefore, represents a promising first step towards the large scale optimization of 3L-CVRP.

This work makes three main contributions to the literature. 
First, it merges the reinforcement learning research streams on bin-packing and vehicle routing by proposing the first reinforcement learning model to solve the 3L-CVRP in literature.
Second, it addresses the computational limitations of existing approaches with the favourable compute times of reinforcement learning during execution.  
Third, the open-sourcing of the reinforcement learning environment lowers the entry barrier for other researchers to extend the literature on 3L-CVRP.

The remainder of this work is structured as follows. In Section \ref{sec:related}, we review the literature and reveal shortcomings of current methods and gaps in the literature. In Section \ref{sec:Model}, we define the problem formulation and develop our reinforcement learning model. In Section \ref{sec:results}, we benchmark the model against state-of-the-art solutions and interpret the results. In Section \ref{sec:discussion}, we discuss the contributions to literature, limitations, and implications for practice before concluding in Section \ref{sec:conclusion}.

\section{Related Work}\label{sec:related}

Two streams of research are particularly relevant for this work: First, the literature on vehicle routing and bin-packing problems. Second, the applications of reinforcement learning in operations management.  

\subsection{Vehicle routing and bin-packing problems}

Increasing the utilization of HGVs requires solving two interdependent optimizations problems: vehicle routing and bin-packaging.
The vehicle routing problem itself focuses on finding the shortest possible vehicle routes to fulfill customer demands \citep{laporte2013vehicle} and is closely related to the traveling salesman problem \citep{flood1956traveling}. However, the problem formulation does not consider the loading situation of vehicles and is therefore not satisfactory for real-world problems.
The bin-packing problem on the other hand solely focuses on packing items as efficiently as possible with no regard to routing considerations \citep{mack2012heuristic}. Therefore, we require the joint solving of both problems in practice.

To combine the vehicle routing and bin-packing problems into one, \citet{gendreau_tabu_2006} introduced the capacitated vehicle routing problem with 3D loading constraints. The \citet{gendreau_tabu_2006} formulation of the 3L-CVRP was originally solved via a tabu search approach by the authors. As surveyed by \citet{bortfeldt_split_2020}, numerous heuristic and metaheuristic approaches have been proposed to solve the \citet{gendreau_tabu_2006} formulation with greater speed and smaller optimality gaps. Recent publications include evolutionary local search \citep{zhang_evolutionary_2015}, column generation technique based heuristics \citep{mahvashcolumn}, and adaptive variable neighborhood search \citep{weiadaptive}. There have also been other formulations of the 3L-CVRP including \citet{bortfeldt_split_2020}, \citet{zhang_hybrid_2017} and \citet{pan_hybrid_2021} who have worked on more realistic 3L-CVRP problems that include time constraints and split-deliveries.

All existing methods share two limitations. First, heuristics-based approaches are naturally only as good as the chosen rules and therefore need to be adapted to new scenarios that deviate from the original problem setting. Second, the existing methods are based on incremental convergence towards a near-optimal solution. This leads to significant compute times as the problem size increases and therefore requires regional restriction of the optimization to compute results on a day-to-day basis. We address these shortcomings by proposing the use of reinforcement learning to solve the \citet{gendreau_tabu_2006} formulation of the 3L-CVRP.

\subsection{Reinforcement learning in operations management}

Reinforcement learning is a subfield of machine learning in which an agent interacts with an environment while trying to maximize a given reward function. At every time step, the agent receives an observation from the environment and then chooses the best action out of all available actions based on what it has learned so far. This choice is then passed to the environment which calculates the new state after taking the chosen action. The observation of this new state is then passed back to the agent to choose the next action and the process is repeated until the interaction ends. By balancing exploration of uncertain actions and exploitation of known good actions, a decision policy is learned \citep{sutton2018reinforcement}. While the training process is time-intensive, the learned policies only need one compute step per chosen action compared to the previously discussed convergence-based operations research methods. This provides considerable opportunities to speed up solution-finding.

Within operations management, reinforcement learning has already been successfully applied to various inventory models. \citet{oroojlooyjadid2021deep} used reinforcement learning to solve the problems of compute time and the lack of one known ideal policy within the supply chain beer game. They were able to achieve near-optimal results on real-world data and  performed the computations in real-time. \citet{CanDeepR6:online} compared the performance of deep reinforcement learning models to state-of-the-art heuristics for lost sales inventory models, dual sourcing inventory models, and multi-echelon inventory models. Their results matched the performance of state-of-the-art heuristics.

The two sub-problems of 3L-CVRP, vehicle routing and bin-packing, have already been studied independently in the reinforcement learning literature. \citet{peng_deep_2020}, \citet{xin_multi-decoder_2020}, \citet{nazari_reinforcement_2018}, and \citet{kool_attention_2019} demonstrated the viability of reinforcement learning for the vehicle routing problem using encoder-decoder architectures. \citet{duan_multi-task_2019} have demonstrated the viability of reinforcement learning for the bin-packing problem with A/B tests at supermarket warehouse systems of Taobao, where they achieved an average of 5.47\% cost reduction.
However, to the best of our knowledge, the 3L-CVRP has not been solved with reinforcement learning in the literature. This work aims at closing this gap by combining vehicle routing and bin-packing into one model.

\section{3L-CVRP Model}\label{sec:Model}

This section develops a reinforcement learning model to solve the 3L-CVRP. In the following, we provide a problem definition and the model specification. 

\subsection{Problem Definition}

While there are numerous formulations of the 3L-CVRP  \citep[e.g.,][]{gendreau_tabu_2006, zhang_hybrid_2017, pan_hybrid_2021}, they generally consist of a graph $G(V,E)$ with vertices $V=\{ 0, 1, ..., n\}$ and edges $E=\{ 0, 1, ...\}$. The vertices $V$ consist of $n$ clients where vertex 0 corresponds to the depot from which all packages originate. The number of edges corresponds to the total number of trips from one location to another by the vehicles.
Each of the $n$ clients is associated with $k=\{1, ..., m_i\}$ packages $\mathcal{I}_{ik}$.  Hereby, the index $i$ represents the client and $k$ represents the package number (e.g., $\mathcal{I}_{12}$ is the second package of client one).
The individual packages have a weight $d_{ik}$, height $h_{ik}$, width $w_{ik}$, and length $l_{ik}$. To deliver the packages, there are $v$ homogeneous vehicles $\mathcal{V}_i$  with a loading space of volume $h_{veh} * w_{veh} * l_{veh}$ with a weight capacity $d_{veh}$. The package placement in each vehicle must fulfill preset constraints (e.g., unloading order of packages). The total cost of the instance is calculated by summing up the cost of all edges $e_{ij}$ (e.g., euclidean distance between two vertices). Hereby, $e_{01}$ would represent a trip from the depot to client one. As defined by \citet{zhang_hybrid_2017}, our objective is therefore to minimize the total travel distance of all vehicles, with $t_i$ being the number of vertices in the route of vehicle $i$, resulting in

\begin{equation}
min\sum_{i=1}^{v}\Big(\sum_{j=1}^{t_i-1} e_{j(j+1)}\Big).
\end{equation}

\subsection{Model Specification}

We base our model on the attention model introduced by \citet{kool_attention_2019} and extend it to solve the 3L-CVRP. The encoder-decoder architecture by \citet{kool_attention_2019} was chosen as the starting point due to its demonstrated generalization performance on various routing problem settings ranging from the orienteering problem to vehicle routing.
Due to the container loading constraints in $n=3$ dimensions, the feasibility checks for $m$ package placements require significant compute time and scale with $O(m * n^3)$. In order to reduce the number of placement locations to check, we use a least-space-wasted-furthest-back-rightmost-lowest package placement heuristic. Such a combination of reinforcement learning with heuristics-based placement has already been used \cite{hu_solving_2017} for the 3D bin-packing problem and outperformed purely heuristics-based approaches.

\subsubsection{Encoder}

\begin{figure}[h]
  \centering
  \includegraphics[width=0.9\textwidth]{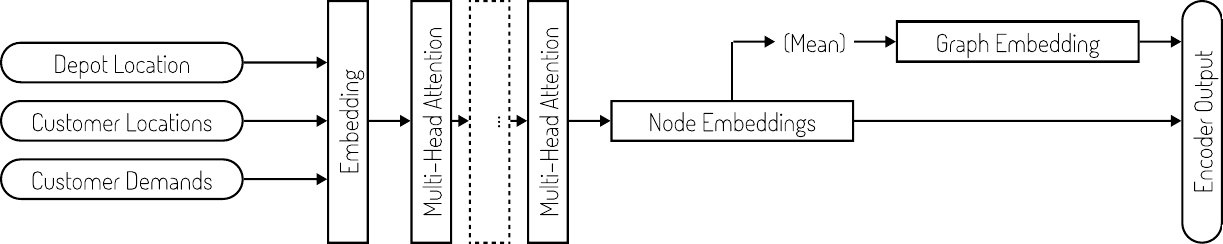}
  \caption{Encoder}
  \label{fig:enc}
\end{figure}

Adapted from \citet{kool_attention_2019}, the input of the encoder shown in Figure \ref{fig:enc} consists of the depot location, the customer locations, and the customer demand. The depot location is defined by a tensor of shape [batch size, 1, 2] that describes the x and y coordinates of the depot. Similarly, the client locations are defined by a tensor of shape [batch size, $n$, 2] that describes the x and y coordinates for each node $n$. The customer demand tensor is of shape [batch size, $\sum_{i=1}^{n} m_i$, 5] and describes the height, weight, length, fragility, and weight of each individual package.

In order to make the model agnostic to vehicle and package sizes, we scale all dimensions relative to the vehicle dimensions (e.g., $h_{veh\text{\_} scaled} = h_{veh}/h_{veh} = 1$ and $h_{ik\text{\_}scaled} = h_{ik}/h_{veh}$). This information is then embedded in $d_h$ dimensions and passed through $a$ multi-head attention layers with $l$ heads each. The multi-head attention allows the model to focus on multiple information criteria of interest at the same time (e.g., dimensions, distances). The generated node embeddings, denoted by a tensor of shape [batch size, $n$, $d_h$], are additionally aggregated as a graph embedding of shape [batch size, 1, $d_h$] and then passed to the decoder as input.

\subsubsection{Decoder}

\begin{figure}[h]
  \centering
  \includegraphics[width=0.9\textwidth]{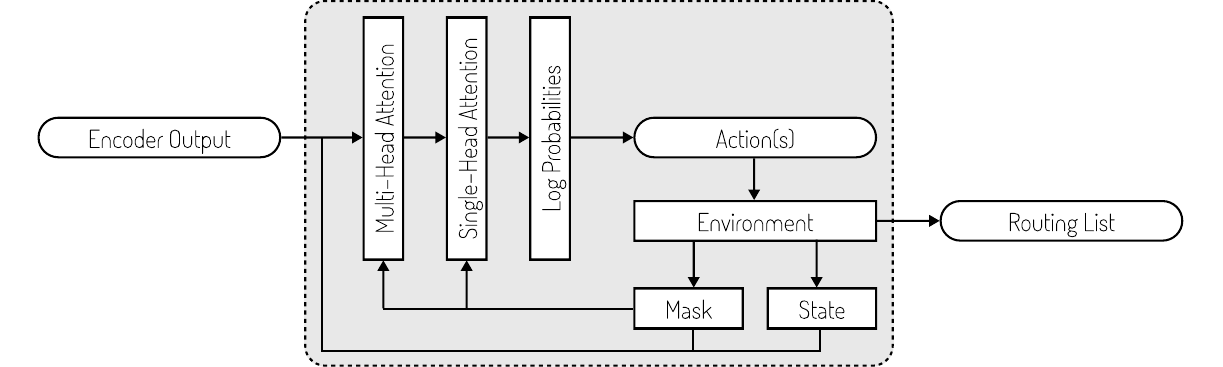}
  \caption{Decoder}
  \label{fig:dec}
\end{figure}

The decoder shown in Figure \ref{fig:dec} combines the initial problem setting information from the encoder with the current loading and packing state to select the most suitable package to be loaded next. After loading the selected package, the current state is updated and the process is repeated until all packages have been loaded.

\begin{figure}[h]
  \centering
  \subfloat[3D container representation]{\includegraphics[width=0.6\textwidth]{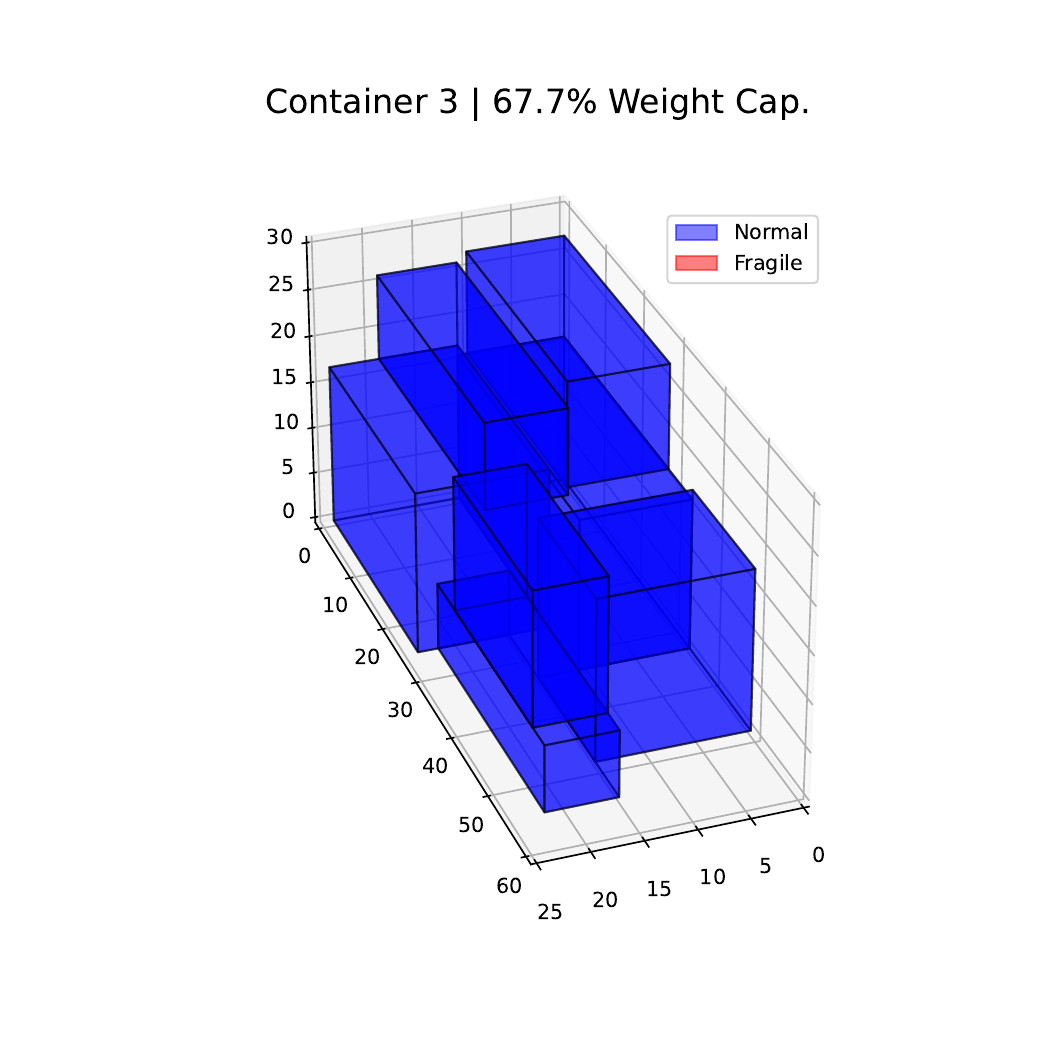}\label{fig:fc1}}
  \hfill
  \subfloat[2D container representation]{\includegraphics[width=0.40\textwidth]{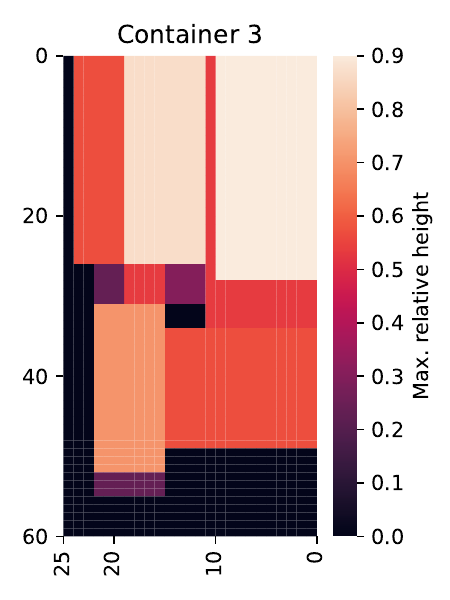}\label{fig:fc2}}
  \caption{Container representation}
\end{figure}

Similarly to \citet{kool_attention_2019}, our step context information consists of three inputs. First, we extract the encoder embedding of the previously chosen node to represent our vehicle routing location, denoted by a tensor of shape [batch size, 1, $d_h$]. Second, we create an embedding of the remaining weight capacity of the current vehicle, also denoted by a tensor of shape [batch size, 1, $d_h$]. Third, we create an embedding of the current 3D container loading state. For computational efficiency, we translate the 3D loading space shown in Figure \ref{fig:fc1} into a 2D representation shown in Figure \ref{fig:fc2} and represent the loading heights via the cell value \citep{zhao2020online}. 
This representation still contains the same information as the 3D representation with improved efficiency for usage with convolutional neural networks (CNNs). In order to represent the fragility of packages, we assign a positive sign to all non-fragile height aggregations and a negative sign to all fragile ones. All values are then scaled from a $[-h_{veh}, h_{veh}]$ to $[-1,1]$ range before resizing the $w_{veh}*l_{veh}$ vehicle representation to a uniform size of $w_{cnn} * l_{cnn} $. This enables the model to be applied to varying container sizes. The scaled and resized 2D input is passed to a 2D convolution layer and subsequently embedded in $d_h$ dimensions. The resulting embedding is of the shape [batch size, 1, $d_h$] and concatenated with the two other step context embeddings.

The encoder output along with the step context embeddings and a mask is passed to a multi-head attention layer which produces an output of the shape [batch size, 1, $d_{h}$]. The mask filters out infeasible actions during calculation (e.g., packages that have already been packed). A single-head attention layer is used to generate a tensor of the shape [batch size, 1, actions] that represents the log probabilities of all actions \citep{kool_attention_2019}. The actions are then passed to the environment to be updated (i.e. loading and routing the selected package) and the decoder step is repeated until all packages have been packed or no more viable placement options remain. The collection of all chosen actions is saved as the routing list.

\subsubsection{Masking and Packing Heuristic}

As computational speed is one of the main objectives of applying reinforcement learning to the 3L-CVRP, we do not compute all package placement feasibilities at every step since they are computationally expensive. Instead, we use a two-step masking approach. The first mask is determined by violations of the problem constraints that do not require a placement feasibility check (e.g., packages that are already packed) and is applied before calculating the action probabilities. 
Rather than passing a single action from the calculated probabilities shown in Figure \ref{fig:dec}, we pass a list of recommended actions in descending order to the environment. This action list is then used for our on-demand look-ahead feasibility check. Starting from the action with the highest probability we perform a placement feasibility check until we find an action that is feasible to be placed in the vehicle based on the constraints (see Algorithm 1 in the Appendix). This way we can reduce computation time significantly as we only have to check one package in an ideal case and all packages in the worst case. To speed up learning of the \citet{gendreau_tabu_2006} constraint that demands that all packages for a location must be delivered in a single vehicle, we add a look-ahead feasibility check. This means when there is more than one package remaining for a location, we check if they are all feasible to be loaded in the current vehicle.

Analogously to \citet{hu_solving_2017}, we place packages in the lowest, rightmost, and furthest back position available in each vehicle to reduce the action space with minimal impact on optimality. This is achieved via a least-space-wasted-furthest-back-rightmost-lowest heuristic. When an action is passed to the feasibility check outlined in Algorithm 1, the corresponding package is selected and the furthest back rightmost lowest placement is computed for all rotation possibilities. The rotation and placement with the least space wasted is selected, with the shorter total loading distance acting as a tiebreaker. Least space wasted is defined as the empty space blocked behind a package along the unloading axis.

\subsubsection{Proximal Policy Optimization with Greedy Rollout}\label{sec:ppo}

Given the complexity and a high number of constraints for our model, we use proximal policy optimization (PPO) to stabilize the learning process \citep{schulman2017proximal}. Our loss function is defined as

\begin{equation}
L(\theta) = \hat{\mathbb{E}}_t[min(r_t(\theta) \times \hat{A}_t, clip(r_t(\theta), 1- \epsilon, 1+\epsilon) \times \hat{A}_t) + \alpha \times S[\pi_\theta](s_t)].
\end{equation}

Here, $S[\pi_{\theta}]$ denotes the entropy of the chosen actions and is used to encourage exploration. The ratio $r_{t}(\theta) = \frac{\pi_{\theta}(a_{t}|s_{t})}{\pi_{\theta_{old}}(a_{t}|s_{t})}$
reflects the change of probabilities in between policy updates. Combined with the clipping parameter $\epsilon$, this ensures model updates that stay close to the original policy. The \textit{advantage} $\hat{A}_t$ is defined as the difference between the cost and a baseline. Similar to \citet{kool_attention_2019}, we use a greedy rollout baseline to compare the current cost of the training epoch to the performance of the previously best model during training. We hereby always select the action with the highest probability for the baseline---thus greedy rollout (in comparison to sampling based on the probability during training). This direct comparison provides a robust low-variance advantage estimation. Our cost function (the negative reward function), is defined as:

\begin{equation}\label{eq:C}
C = C_{vrp} + C_{packing}
\end{equation}

\begin{equation}\label{eq:vrp}
C_{vrp} =\frac{\sum_{i=1}^{v}(\sum_{j=1}^{t_i-1} e_{j(j+1)})}{p_{vrp}*\sum_{i=1}^{n} e_{i(j=0)}}
\end{equation}

\begin{equation}\label{eq:pack}
C_{packing} = \frac{\text{number of missed packages}}{n}
\end{equation}

Equation \ref{eq:vrp} sums up all vehicle paths and divides them by the sum of all individual travel distances from the depot to each of the $n$ client locations multiplied by a penalty factor $p_{vrp}$. This scaling ensures that varying optimal distances caused by the randomly generated locations are comparable during training.
As our agent would learn to not load any packages for a distance of zero, we penalize packages that were not loaded in Equation \ref{eq:pack}. The division by $n$ is used to keep both cost function terms similar in magnitude. 
Our formulation enables the weighing of the vehicle routing ($p_{vrp} \to 0$)  and bin-packing  ($p_{vrp} \to \infty$) aspects during training.

 \section{Results}\label{sec:results}
 
 This section reports on the implementation details and the results of the proposed RL model on the benchmarking instances of \citet{gendreau_tabu_2006}.
 We compare against the original tabu-search of \citet{gendreau_tabu_2006}, the column-generation technique of \citet{mahvashcolumn}, and the evolutionary local search of \citet{zhang_evolutionary_2015} in terms of the achieved routing distances and the time to compute the solutions.

\subsection{Implementation Details}

This subsection focuses on the problem definition and the training process. We describe the used 3L-CVRP problem formulation, the chosen benchmarking instances, and the training process including data generation.

\subsubsection{Benchmarking Problem Definition}\label{chap:def}

\citet{gendreau_tabu_2006} define the cost of travelling from one vertex to another as the euclidean distance between the two and set the following constraints:

\begin{enumerate}
    \setlength{\itemsep}{0pt}
    \setlength{\parskip}{0pt}
    \item Every vehicle route must start and end at the depot.
    \item Clients may not be visited more than once.
    \item All package weights $d_i$ combined in a vehicle must not exceed $d_{max}$.
    \item All packages must be loaded in an orthogonal three-dimensional layout without overlapping and within the vehicle space.
    \item The vertical orientation of packages is fixed.
    \item Fragile packages $f_{ik}=1$ can be placed on non-fragile packages $f_{ik}=0$ but not vice-versa.
    \item Each package must be supported by either the floor of the vehicle or other packages. The created support area $a_{supp}$ must fulfil $a_{supp} >= a_{min} * w_{ik}*l_{ik}$ with $a_{min}$ being the support area threshold.
    \item All packages must be possible to unload via the Last In First Out (LIFO) principle along the $l$ axis of the vehicle without intersecting other packages or decreasing the support area of other packages.
\end{enumerate}

\subsubsection{Benchmarking Instances}

\citet{gendreau_tabu_2006} introduced 27 predefined instances of varying complexity based on the problem formulation described in Section \ref{chap:def}. Within the scope of this work, we focus on the two smallest instances, as shown in Table \ref{tab:inst}, due to the significant training time required for bigger instances.

\begin{table}[h]
\caption{Predefined \citet{gendreau_tabu_2006} instances}
\centering
\begin{tabular}{llll}
\toprule
Instance         & Destinations & Vehicles & Packages \\
\midrule
E016\text{-}03m & 15               & 5            & 32           \\
E016\text{-}05m & 15               & 5            & 26          \\
\bottomrule
\end{tabular}
\label{tab:inst}
\end{table}

\subsubsection{Training Process}\label{sec:train}

The training process for our reinforcement learning model is as follows. At each epoch, we generate 100 random instances which are processed in parallel. This reduces the variance during updates after each epoch while still being fast enough for training on our hardware (ca. 60 seconds per epoch on the ETH Zürich Euler cluster with 8 cores). 
We generate the random instances by uniformly sampling from the parameters listed in Table \ref{tab:par}. These ranges are chosen to resemble the ranges used by \citet{gendreau_tabu_2006} for the design and validation of their own tabu-search method. 
We speed up training by using training instances with fewer size increments due to a smaller container size than those used in the benchmarks (e.g., width dimension reduction from 25 to 5). This reduces the compute time for feasibility checks in the containers significantly as the search space shrinks by a factor of 125 from $30*25*60=45.000$ to $6*5*12=360$.

\begin{table}[h]
\caption{Parameter space for training instances}
\centering
\begin{tabular}{llll}
\toprule
Parameter         & Value(s) \\
\midrule
$n$ & 15             \\
Probability of package being fragile & 25\%         \\
$x_{i}$ & [0, 100]            \\
$y_{i}$ & [0, 100]           \\
$h_{veh}$ & 6             \\
$w_{veh}$ & 5            \\
$l_{veh}$ & 12            \\
$h_{ik}$ & [$0.2 \times h_{veh}$ , $0.6 \times  h_{veh}$] \\
$w_{ik}$ & [$0.2 \times w_{veh}$ , $0.6 \times  w_{veh}$] \\
$l_{ik}$ & [$0.2 \times l_{veh}$ , $0.6 \times  l_{veh}$] \\
$m_i$ & [1, 2, 3]          \\
$d_i$ & [1, 2, ..., 30]          \\

\bottomrule
\end{tabular}
\label{tab:par}
\end{table}

Based on the total volume and weight of all packages, we assign a number of vehicles to each instance that equals twice the needed capacity. This ensures that all packages can be packed without the computationally intensive step of testing for the minimum number of trucks based on feasible loading combinations. This upper truck limit prevents infinite looping of extra requested trucks during training.

\begin{figure}[h]
  \centering
  \subfloat[Average validation cost of runs]{\includegraphics[width=0.5\textwidth]{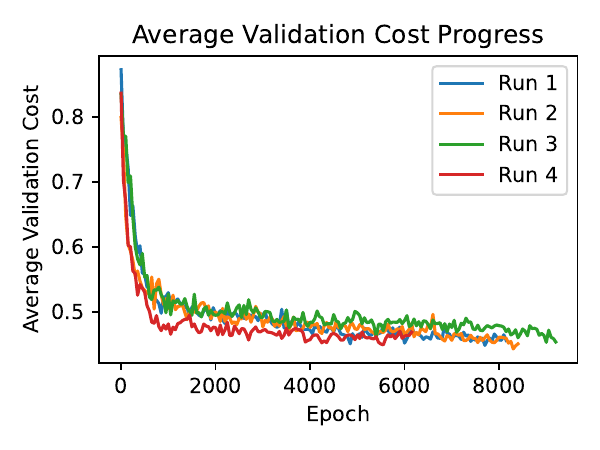}}
  \hfill
  \subfloat[Average missed packages of runs]{\includegraphics[width=0.5\textwidth]{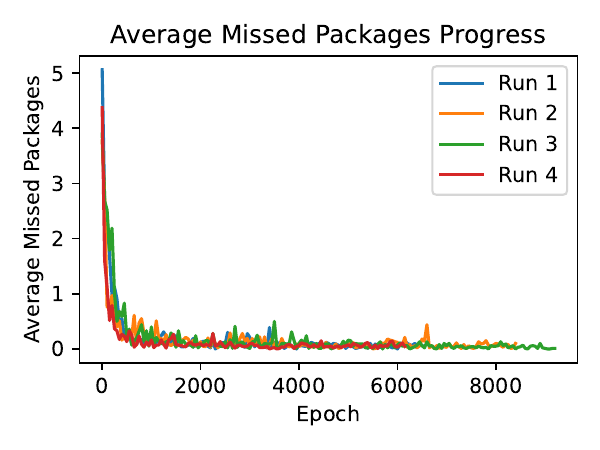}}
  \caption{Validation data learning curves}
  \label{fig:learn}
\end{figure}

After each epoch, the cost and loss for the routing is calculated according to Section \ref{sec:ppo}. Next, the encoder and decoder are optimized with the Adam optimizer \citep{kingma2014adam} before generating new data for the following epoch. Figure \ref{fig:learn} shows the learning curve of our model for four 120 hour runs with different random seeds on the validation data (batch size equals training). Run 4 was trained with a batch size of 128 in comparison to all other runs with size 100 and thus performed fewer epochs within 120 hours. The model parameters are listed in Table \ref{tab:paras} in the appendix. We selected the epoch with the best validation performance to be used for the benchmarking on the \citet{gendreau_tabu_2006} instances in Section \ref{sec:bench}.

\subsection{Benchmarking Results}\label{sec:bench}

In this subsection, we compare the performance of our reinforcement learning model against state-of-the-art methods for solving the 3L-CVRP. We compare both the routing results as well as the compute times and interpret the results.

\subsubsection{Routing Distances}\label{sec:dist}

 We compare our results to three existing approaches. First, the original tabu-search implementation by \citet{gendreau_tabu_2006} to illustrate the performance improvement of recent publications. Second, the evolutionary local search implementation of \citet{zhang_evolutionary_2015} which provides the best results in terms of pure routing distance. Third, the column-generation based approach of \citet{mahvashcolumn} because of its compute speed (33.7\% faster than \citet{zhang_evolutionary_2015}) paired with only a slight decrease in routing performance. Given that a good balance between routing distance and compute time is desirable for real-world applications, we will refer to \citet{mahvashcolumn} as the state-of-the-art to beat.
 
 \begin{table}[h]
\caption{Routing distance comparison of models}
\centering
\begin{tabular}{llll}
\toprule
         & E016\text{-}03m & E016\text{-}05m & Avg. distance   \\
         \midrule
\citet{gendreau_tabu_2006} & 316.32   & 350.58   & 333.45 \\

\citet{zhang_evolutionary_2015}        & 302.02   & 334.96   & 318.49\\
\citet{mahvashcolumn}        & 315.16   & 345.28   & 330.22 \\

\midrule
\textbf{RL (our best model from run 2)}     & \textbf{337.85}   & \textbf{347.86}   & \textbf{342.86} \\
RL (average of 4 runs)     & 358.26   & 356.80   & 357.53 \\

\midrule
Gap to \citet{gendreau_tabu_2006}     & 6.81\%  & -0.78\%   & 2.82\% \\

Gap to \citet{zhang_evolutionary_2015}     & 11.86\%  & 3.85\%   & 7.65\% \\
\textbf{Gap to \citet{mahvashcolumn}}     & 7.20\%  & 0.75\%   & \textbf{3.83\%} \\
\bottomrule
\end{tabular}
\label{tab:gend}
\end{table}

\begin{figure}[h]
  \centering
  \subfloat[E016\text{-}03m instance]{\includegraphics[width=0.5\textwidth]{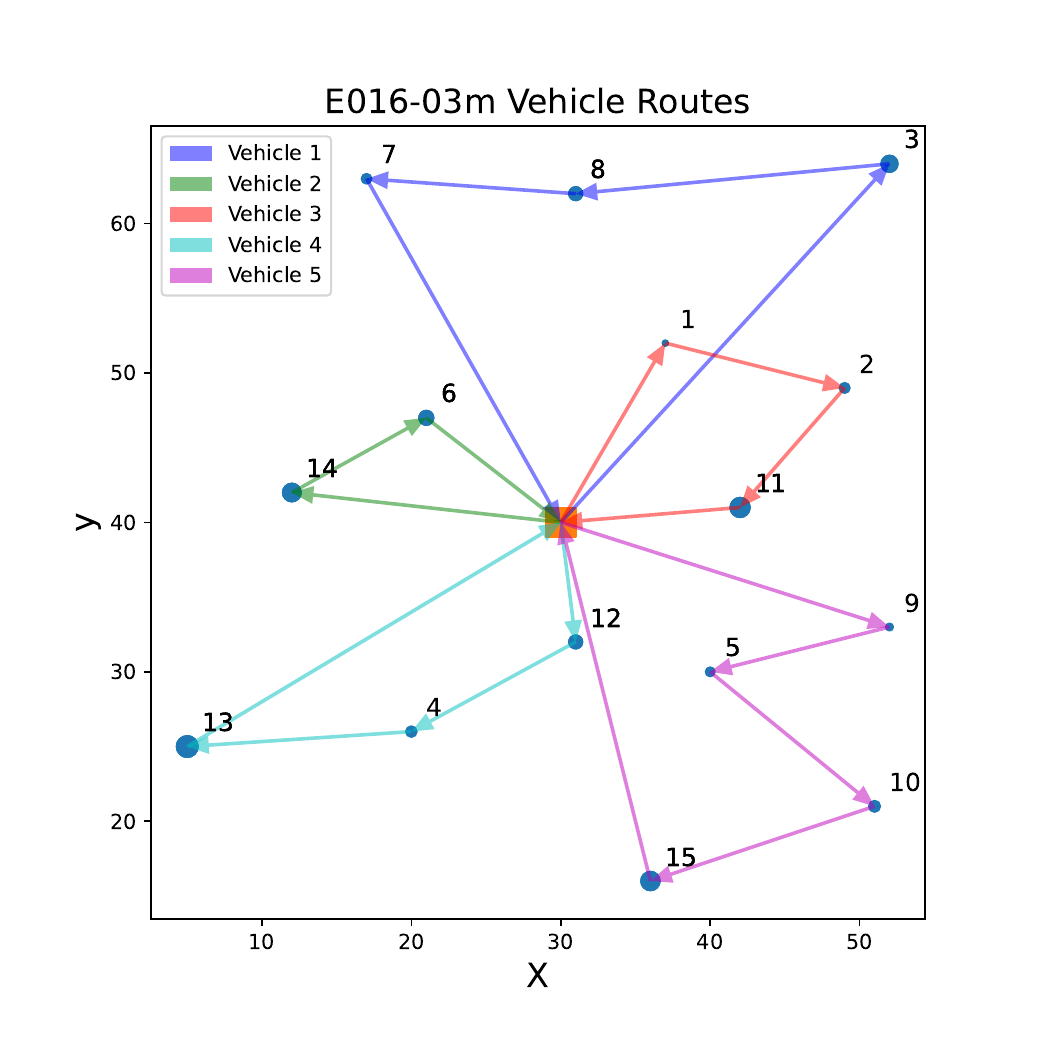}\label{fig:f1}}
  \hfill
  \subfloat[E016\text{-}05m instance]{\includegraphics[width=0.5\textwidth]{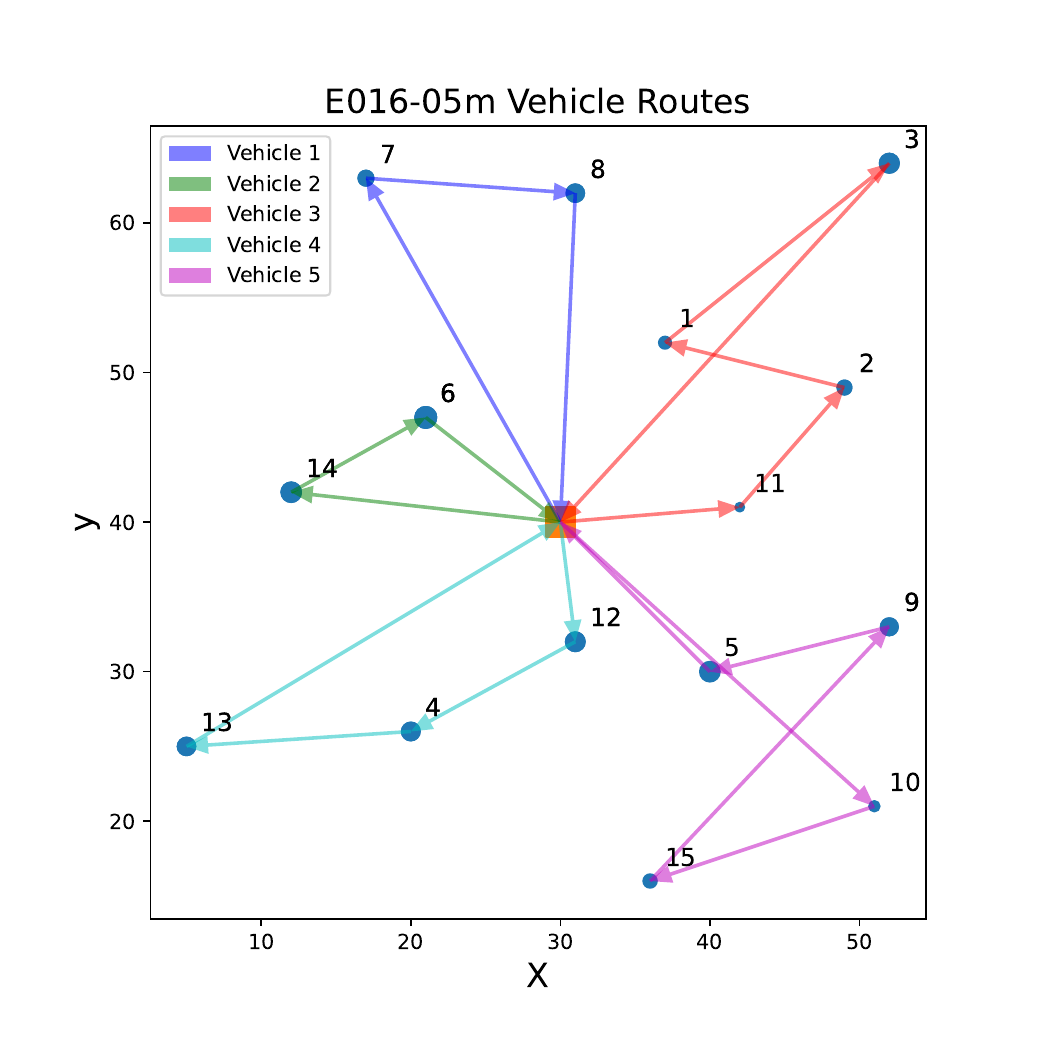}\label{fig:f2}}
  \caption{Vehicle routing with our model}
\end{figure}

 Table \ref{tab:gend} shows the achieved routing distances of our model compared to the literature. Our model trained in run 2 achieved the best results on our validation data as well as the \citet{gendreau_tabu_2006} instances. The vehicle routing and packing by this model can be seen in Figure \ref{fig:f1} and \ref{fig:f2}. On the E016\text{-}03m instance our model achieves a routing distance of 337.85. Compared with the solutions from the literature this is a  6.81\% to 11.86\% gap. On the E016\text{-}05m instance our model achieves a routing distance of 347.86 with a gap ranging from -0.78\% to 3.85\%.

In order to test the robustness of our results, we perform test-time augmentation for the benchmarking instances. Hereby, we translate and flip the node locations of the two benchmarking instances to investigate how this affects the outputs. Given the [0, 100] plane we are using, flipping means that $x=10$ would become $x=90$ and vice versa. Given that the instances are unchanged except for translation or mirroring, the routing should remain the same in an ideal model. The results in Table\ \ref{tab:tta} show that the two best-performing runs on the validation data do not produce consistent results during test-time augmentation. Run 4, which has been trained with a batch size of 128 achieves consistent results when all nodes are shifted along the x and y-axis. The model of run 2 with a smaller batch size of 100 is not consistent for a larger movement of +20. Both models produce different results when an axis is flipped. Thus, steps need to be taken to improve the robustness of our models (e.g., train-time augmentation, increased batch sizes for model updates with less variance) to produce robust results.

\begin{table}[h]
\caption{Test-time augmentation}
\centering
\begin{tabular}{llllll}
\toprule
 & E016\text{-}03m & &  E016\text{-}05m &  \\
\cmidrule(r){2-5}

                     & Run 4 & Run 2 & Run 4 & Run 2 \\
\midrule
Original             & 334.99   &            337.85            & 361.86  &      347.86                   \\
x and y + 10         & 334.99     &          337.85            & 361.86  &      334.95                   \\
x and y + 20         & 334.99    &            303.51*          & 355.04  &  358.80 \\ 

x-axis flipped       & 362.81    &         329.85*              & 347.06   & 364.19                       \\
y-axis flipped       & 330.54*  &      345.78           & 365.84   & 348.38                       \\
x and y-axis flipped & 380.53       &       359.81             & 328.38  &  355.25*                       \\

                 \bottomrule 
                \multicolumn{5}{r}{* not all packages packed}
\end{tabular}
\label{tab:tta}
\end{table}

  As mentioned in Section \ref{sec:train}, our model was trained on a different container size ([6,5,12]) than the benchmarks ([30,25,60]). Achieving results within a few percent of the state-of-the-art solution shows that our model is container size agnostic and can handle package sizes it has never encountered before. The current 3.83\% gap in routing distance compared to \citet{mahvashcolumn}, therefore, seems achievable by improving the model architecture (e.g., improving the placement heuristic), training on the container size used for benchmarking itself and increasing the batch size for more robust results.

\subsection{Compute Times}

Next, we compare the compute time scaling of our best performing model against the compute times reported by \citet{mahvashcolumn}.  \citet{mahvashcolumn} was chosen as a baseline because of their 33.7\% faster computation of solutions compared to \citet{zhang_evolutionary_2015}. As absolute compute time comparisons can vary greatly by the chosen implementation and hardware (e.g., C, Python, GPU, CPU), we focus on the relative scaling of computing times with increasing instance sizes. 

\begin{figure}[h]
  \centering
  \includegraphics[width=0.6\textwidth]{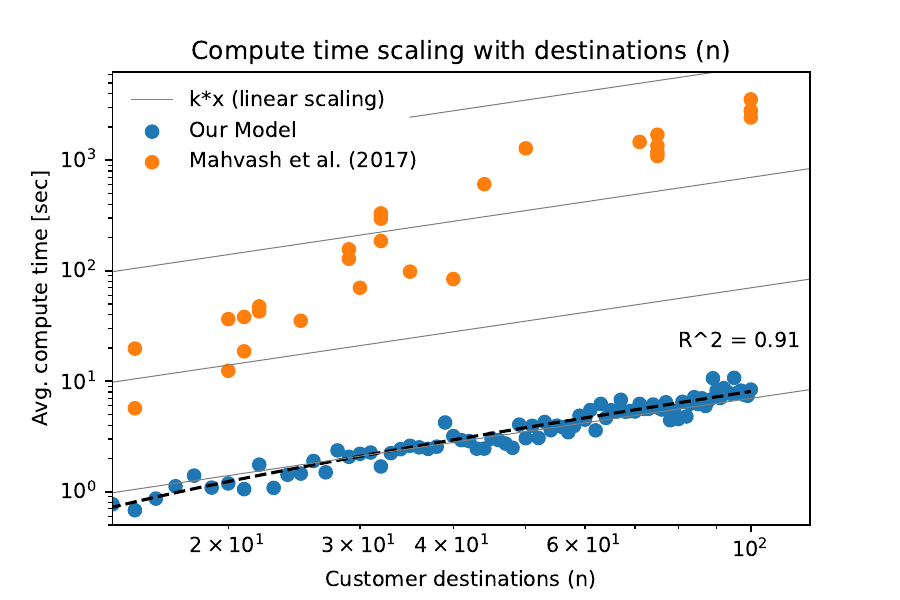}
  \caption{Comparison of compute time scaling between our model on randomly generated instances with \citet{gendreau_tabu_2006} constraints and \citet{mahvashcolumn} on the 27 \citet{gendreau_tabu_2006} instances}
  \label{fig:scale}
\end{figure}

The compute times for our model in this comparison were calculated by applying our model to randomly generated instances according to the \citet{gendreau_tabu_2006} formulation of the 3L-CVRP. The compute times for \citet{mahvashcolumn} are the ones they reported for the solution of the respective \citet{gendreau_tabu_2006} benchmarking instances. Thus, the absolute comparison due to varying hardware should not be considered and the focus should be on the scaling.
Figure \ref{fig:scale} shows empirically that our model scales approximately linearly ($R^2=0.91$) with an increasing number of destinations. The column-generation technique by \citet{mahvashcolumn} on the other hand does not scale linearly. 

Figure \ref{fig:scale} highlights the potential of our model to solve the shortcoming of existing methods in regard to compute time scalability. This enables the extension of the real-life solution space beyond current regional limits as discussed in Section \ref{sec:intro} and will allow for solutions closer to a global optimum rather than regional optima.

\section{Discussion}\label{sec:discussion}

In this section, we discuss our contribution to the literature, the limitations of our current model, and the implications for practice.

\subsection{Contributions to the Literature}

This work proposes a reinforcement learning model to solve the 3L-CVRP. Thereby, we contribute to the literature in three major ways.
First, this work is the first to apply reinforcement learning to the 3L-CVRP. We combine the reinforcement learning research streams of vehicle routing and bin-packing and lay the foundation for future work on this problem.
 Second, we address the problem of computational scalability in existing 3L-CVRP research by introducing a method with vastly improved scalability. This lays the foundation for future research into pure reinforcement learning approaches and hybrid approaches for the 3L-CVRP that can be applied to larger scenarios than current methods.
 Third, we will make our reinforcement learning environment open source to make research into this topic more accessible. 

\subsection{Limitations}\label{sec:limits}

 Our proposed reinforcement learning model has three main limitations. 
 First, our current model is designed to meet the \citet{gendreau_tabu_2006} requirements and needs to be adapted for real-world use-cases. This includes among others heterogeneous fleets, package pickups during routing, time constraints for deliveries, and realistic loading constraints.
 Second, the model relies on a placement heuristic. By replacing it with a non-heuristics-based approach, we can gain access to an increased solution space and thus improve routing distance results. While this is a considerable limitation for the \citet{gendreau_tabu_2006} formulation with greatly varying package sizes, we chose the heuristic on purpose due to the alignment with practice. The real-world use-case of our industrial partner is based on standardized pallets within standardized containers which aligns well with the chosen heuristic. The hybrid approach of reinforcement learning and a placement heuristic helps to keep the model complexity and training time lower.
Three, test-time augmentation of the instances shows that our models need to be improved in terms of robustness.

We also acknowledge the limitations of reinforcement learning in general.
 Deep reinforcement learning models in general have two significant limitations that apply to the 3L-CVRP. First, out-of-sample data can produce unreliable results. Therefore, the data for training must be chosen carefully, models must be designed to be robust to outliers, and models must be limited in their usage to the designed application area. Second, the model is a black box and thus may cause stakeholders to be wary of implementing such a model as a first adopter.
 
In order to leverage the full potential of reinforcement learning for the 3L-CVRP and beyond, these limitations must be addressed. In regards to the general reinforcement learning limitations, we aim to integrate recent advances in the fields of robust models \citep{derrow2021eta}, interpretability of reinforcement learning models \citep{madumal_explainable_2020}, and safe exploration \citep{turchetta2020safe} to make the model production-ready. Furthermore, we aim to extend the model to incorporate real-world constraints (e.g., time constraints, pick-up, heterogeneous fleets). 
The improved model will then be applied to real-world data of our industry partner to detect further improvement potential and to estimate the economic impact compared to current solutions.

\subsection{Implications for Practice}

Our results show that reinforcement learning provides promising opportunities to enable large-scale optimization beyond current regional boundaries. In order to increase the likelihood of success, we have two recommendations.
First, we recommend the usage of a hybrid model that combines reinforcement learning with established heuristics similar to our implementation. This combines the scalability of reinforcement learning with the reliability of established heuristics.
Second, we recommend creating different models for easily separable use cases to improve model performance and reliability. Such separable use cases are for example pallets versus cardboard boxes of varying sizes. Another possible split are low-volume areas and high-volume areas (e.g., London-Birmingham-Manchester versus the rest of the United Kingdom). 

\section{Conclusion}\label{sec:conclusion}

This work proposes a reinforcement learning model for the 3L-CVRP to address the main shortcoming, non-linear computational scaling, of current operations research methods. First, we demonstrated the favourable scaling of our model with increasing problem size through computational experiments. Second, we showed that the routing performance of our model lies within 0.75\% to 11.86\% of the current state-of-the-art methods. Based on these findings, we see reinforcement learning as a promising path to large-scale global optimization of logistics operations that lie beyond the computational scope of existing methods and thus unlock new possibilities for efficiency gains and emissions reduction.

\bibliographystyle{agsm}
\bibliography{library}

\newpage
\appendix
\section*{Appendix}
\begin{algorithm}
\caption{Package loading feasibility check for selected container}
\label{alg:euclid}
\begin{algorithmic}[1]
\State $l_{skip}$: prev. determined min. viable $l$

\For{$l$ in range 0 to $l_{container} - l_{package}$}
    \If{$l$ < $l_{skip}$}
        \State skip to next $l$ iteration
    \EndIf
    \State $h_{contour}$ = min. height for each $w$ along $w(l)$
    
    \For{$w$ in range 0 to $w_{container} - l_{package}$}
        \If{$h_{package} - h_{contour}(w) < 0$}
        \State skip to next $w$ iteration
        \EndIf
        
    \For{$h$ in range $h_{contour}(w)$ to $h_{container} - h_{package}$}
        \If{$container(h,w,l)$ not empty}
        \State skip to next $h$ iteration
        \EndIf
        
        \If{$h$ > 0}
        \If{Fragility and min. support area is not fullfilled}
        \State skip to next $w$ iteration
        \EndIf
        \EndIf
        
        \If{LIFO is not fullfilled}
        \State skip to next $h$ iteration
        \EndIf
        
    \State return found feasible location, $(h,w,l)$

    \EndFor
    \If{there were no future feasible placements left}
        \State $l_{skip}$ = $l$
        \EndIf 
    
    \EndFor
   
\EndFor
\State return no feasible location found
\label{alg:feas}
\end{algorithmic}
\end{algorithm}

\begin{table}[]
\centering
\begin{tabular}{ll}
\toprule
Parameter                                                    & Parameter range         \\
\midrule
Batch size (training and validation)                         & 64, \textbf{100}, 128            \\
Learning rate                                                & 1e-3, \textbf{1e-4}              \\
Learning rate decay                                          & \textbf{0.9 per 10000 steps}      \\
Epochs                                                       & \textbf{max. 10000 or 120 hours}  \\
Gradient norm clipping                                       & 0.1, \textbf{0.5}, 1.0           \\
Embedding dimensions $d_h$                                   & \textbf{128}                     \\
Multi-head attention encoder layers $a$                      & \textbf{3}                       \\
Neurons per multi-head attention layer                       & \textbf{512}                     \\
Multi-head attention heads $l$                               & \textbf{8}                       \\
2D resizing target ($w_{cnn}, l_{cnn)}$                           & \textbf{(30, 60)}                  \\
2D convolutional layer kernel size                           & \textbf{(5, 5)}                  \\
Single-head attention tanh clipping                          & \textbf{10}                      \\
Greedy baseline update frequency                             & \textbf{100 epochs}              \\
PPO "minibatches"                                            & 3, \textbf{5}                       \\
PPO clipping $\epsilon$                                      & \textbf{0.2}                     \\
Entropy factor $\alpha$                                      & \textbf{0.0001}                  \\
Penalty factor $p_{vrp}$                                      & 1, \textbf{2}, 10                  \\
\bottomrule
\multicolumn{2}{l}{Notes: The selected parameters are highlighted in bold.}\\
&                    
\end{tabular}
\caption{Training parameters}
\label{tab:paras}
\end{table}

\end{document}